\documentclass[journal]{IEEEtran}

\ifCLASSINFOpdf
\else
   \usepackage[dvips]{graphicx}
\fi
\usepackage{url}
\usepackage{booktabs} 
\usepackage{mathbbol}
\usepackage{tabularx}
\usepackage{adjustbox}
\usepackage{soul}
\usepackage[table]{xcolor}
\usepackage{tabu}
\usepackage{multirow}
\usepackage{makecell}
\usepackage{bbding}
\usepackage{array}
\usepackage{comment}
\usepackage[numbers,sort&compress]{natbib} 
\usepackage{amsmath}
\usepackage{fontawesome} 
\usepackage{amssymb} 

\newcolumntype{I}{!{\vrule width 1pt}}
\newcolumntype{i}{!{\vrule width 1.8pt}}

\usepackage{graphicx}

\begin{document}

\title{GV-VAD : Exploring Video Generation for Weakly-Supervised Video Anomaly Detection }

\author{Suhang Cai*, Xiaohao Peng*, Chong Wang\textsuperscript{\faEnvelope}, Xiaojie Cai, and Jiangbo Qian
\thanks{
Suhang Cai graduated from the Faculty of Electrical Engineering and Computer Science, Ningbo University, Ningbo, Zhejiang 315211, China, and currently works at China Telecom Corporation Limited Wenzhou Branch.  
Chong Wang, Xiaojie Cai, Xiaohao Peng, and Jiangbo Qian are with the Faculty of Electrical Engineering and Computer Science, Ningbo University, Ningbo, Zhejiang 315211, China. E-mail: 2211100245@nbu.edu.cn;
2311100086@nbu.edu.cn; wangchong@nbu.edu.cn; 2211100083@nbu.edu.cn; qianjiangbo@nbu.edu.cn.

*These authors contributed equally to this work and should be considered co-first authors.
\faEnvelope \ Corresponding Author: Chong Wang.
}
}

\markboth{Journal of \LaTeX\ Class Files, Vol. 14, No. 8, August 2015}
{Shell \MakeLowercase{\textit{et al.}}: Bare Demo of IEEEtran.cls for IEEE Journals}
\maketitle

\begin{abstract}
Video anomaly detection (VAD) plays a critical role in public safety applications such as intelligent surveillance.
However, the rarity, unpredictability, and high annotation cost of real-world anomalies make it difficult to scale VAD datasets, which limits the performance and generalization ability of existing models. 
To address this challenge, we propose a generative video-enhanced weakly-supervised video anomaly detection (GV-VAD) framework that leverages text-conditioned video generation models to produce semantically controllable and physically plausible synthetic videos. 
These virtual videos are used to augment training data at low cost. 
In addition, a synthetic sample loss scaling strategy is utilized to control the influence of generated synthetic samples for efficient training.
The experiments show that the proposed framework outperforms state-of-the-art methods on UCF-Crime datasets. The code is available at https://github.com/Sumutan/GV-VAD.git.
\end{abstract}


\vspace{-2mm}

\IEEEpeerreviewmaketitle

\section{Introduction}
\label{sec:intro}

Video anomaly detection (VAD) aims to detect events in videos that significantly deviate from normal patterns. 
Due to its widespread applications in areas such as intelligent surveillance systems and video censorship, VAD has gained increasing attention in both academia and industry.
Most existing VAD methods \cite{
tao2024feature,qian2025ucf,gong2023feature,sultani2018real,tian2021weakly} identify anomalous events by fitting the normal or anomalous visual patterns learned from the training videos. 
Therefore, the quality and quantity of training data play a vital role in determining model performance.
However, existing public datasets exhibit multiple limitations due to the inherent scarcity and unpredictability of real-world anomalies. Firstly, the insufficient quantity and diversity of anomaly samples hinders a comprehensive coverage of real-world scenarios. Moreover, ambiguous boundaries between normal and abnormal behaviors introduce subjective biases during manual labeling. Lastly, data collected from specific environments is hard to support cross-domain detection tasks, constraining model generalization capabilities in practical deployments.

To address the challenge of collecting anomaly videos, UBnormal \cite{UBnormal} explores the effectiveness of constructing synthetic video datasets for video anomaly detection. 
By utilizing the 3D animation and modeling software Cinema4D to create virtual scenes, UBnormal \cite{UBnormal} produces a balanced synthetic dataset with an equitable ratio of normal to anomalous samples. 
The dataset consists of 543 videos, containing 29 virtual scenes with controlled anomalies. 
This synthetic dataset not only facilitates the training and evaluation of video anomaly detection models but also provides a controlled environment to study various anomaly scenarios and their characteristics, thus enabling more robust and generalizable detection methods.

\begin{figure}
    \centering
    \includegraphics[width=\linewidth]{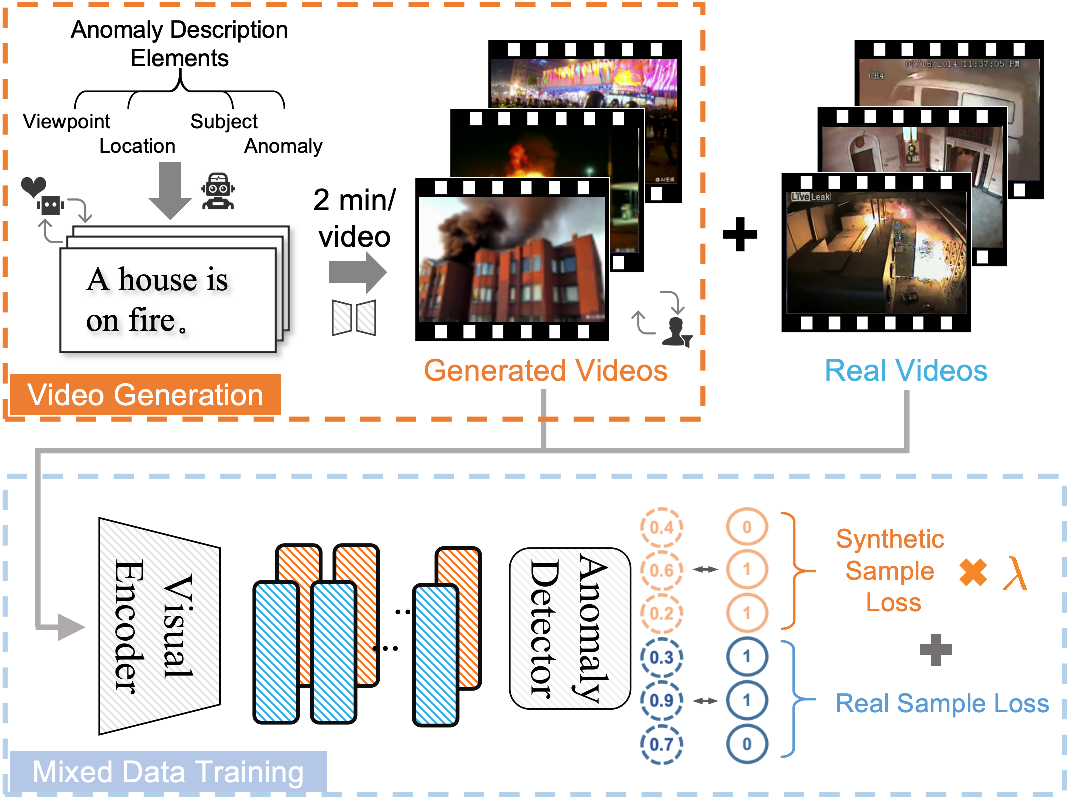}
    \caption{The overview of the proposed generative video-enhanced framework.}
    \label{fig:fig1}
    \vspace{-5mm}
\end{figure}

Although generating virtual videos through 3D modeling software \cite{UBnormal} can mitigate the scarcity of anomaly videos, this approach still requires substantial manual effort and high computational costs (about 15 seconds/frame).
Inspired by the rapid advancement of video generation \cite{Cogvideox}, we propose a generative video-enhanced framework as shown in Fig. \ref{fig:fig1}, which utilizes a diverse set of synthetic anomaly videos generated from desired anomaly description elements.
This framework constructs physically plausible and semantically controllable synthetic videos for effective training data augmentation in video anomaly detection.

\begin{figure*}[t]
    \centering
    \includegraphics[width=\linewidth]{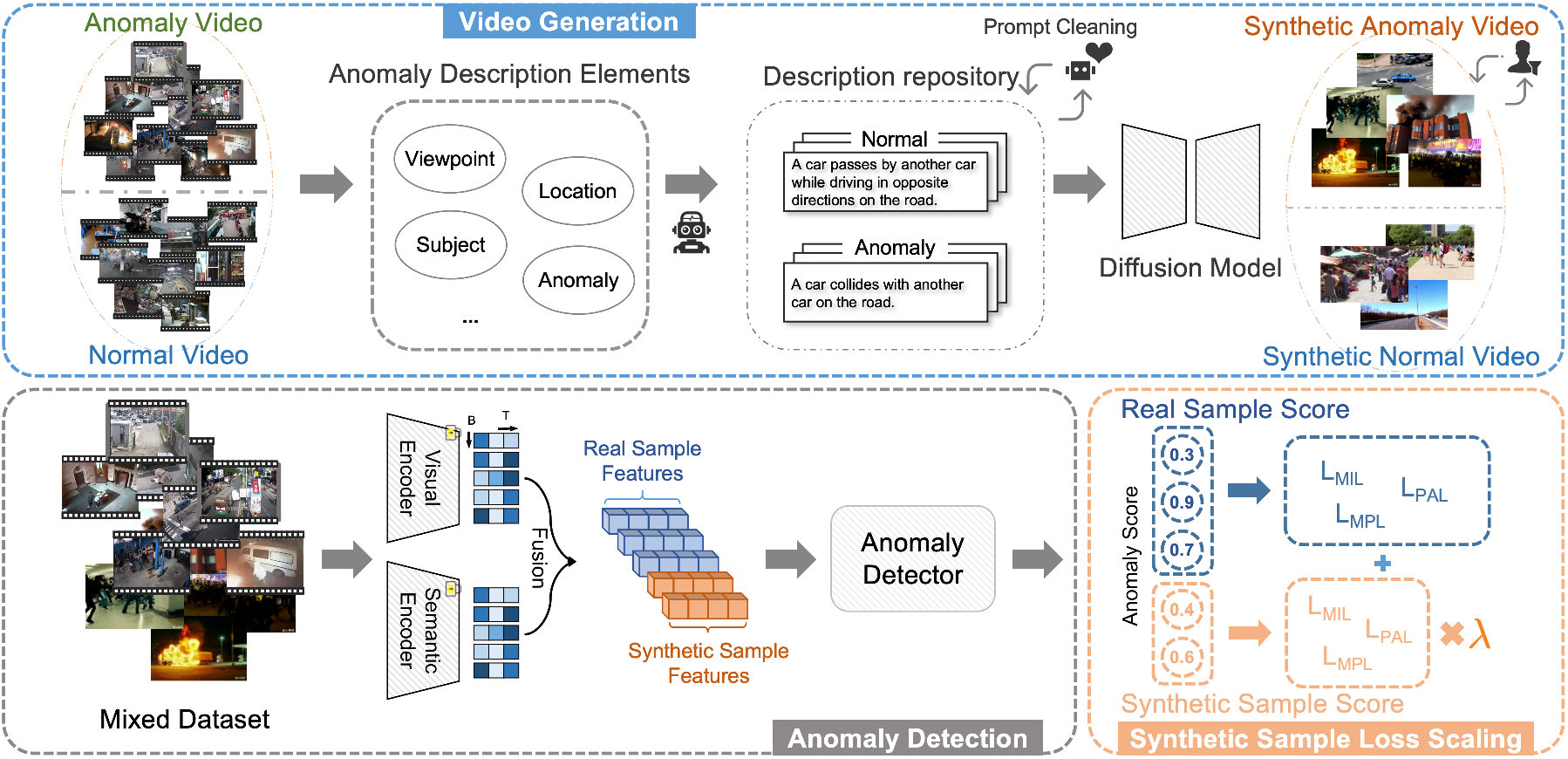}  
    \caption{The overall architecture of proposed GV-VAD framework. 
    The upper part is the illustration of video generation, and the lower is the weakly-supervised learning with synthetic sample loss scalling.Best viewed in color.
    }
    \label{fig:fig2}
\vspace{-3mm}
\end{figure*}

On the other hand, the domain gap has always existed between synthetic videos and real ones. Some works have been proposed to narrow the gap. 
For example, UBnormal \cite{UBnormal} used CycleGAN \cite{cycleGAN} to adjust the style of generated videos, making them more similar to real-world data.
In this work, a synthetic sample loss scaling strategy is proposed to effectively adjust the influence of synthetic samples during the training stage. It prevents the model from overfitting to the virtual data domain while leveraging the pattern and scene information of the virtual data.

\section{Methodology}
\label{sec:Methodology}

To explore the effectiveness of video generation, a new generative video-enhanced weakly-supervised video anomaly detection framework (GV-VAD) is presented. 
Specifically, a conditional diffusion model is employed to generate videos from refined event descriptions, as shown in Fig. \ref{fig:fig2}. 
In addition, a synthetic sample loss scaling is proposed to adjust the influence of synthetic samples on model training.

\subsection{Video Generation}
\label{ssec:Video Generation}


To create sufficient and suitable descriptions for video generation, we analyzed real surveillance videos in current datasets, and identified four core elements that define anomalies, namely camera viewpoint, location, subject, and anomalous event. 
Subsequently, we fed these four elements into a large language model to generate paired semantic descriptions of abnormal and normal events. 
For instance, the input “surveillance camera viewpoint (viewpoint), train station (location), passenger (subject), collapsing (anomalous event)” might generate the anomalous event description: “From the train station surveillance camera, a passenger collapses on the platform, causing brief panic as others rush to help,” and the normal event description: “From the surveillance camera, commuters wait calmly on the platform, reading or checking phones, maintaining a peaceful atmosphere.”

Through this approach, we can create a description repository, which contains anomalous video descriptions 
$\mathcal{C}_{va}=\left \{ \mathcal{C}_{va}^1 , \mathcal{C}_{va}^2 , \mathcal{C}_{va}^m \right \} $ 
and normal video descriptions 
$\mathcal{C}_{vn}=\left \{ \mathcal{C}_{vn}^1 , \mathcal{C}_{vn}^2 , \mathcal{C}_{vn}^m \right \}$.
It includes common anomalous events such as violence, traffic accidents, explosions, etc.
\begin{figure}
    \centering
    \includegraphics[width=\linewidth]{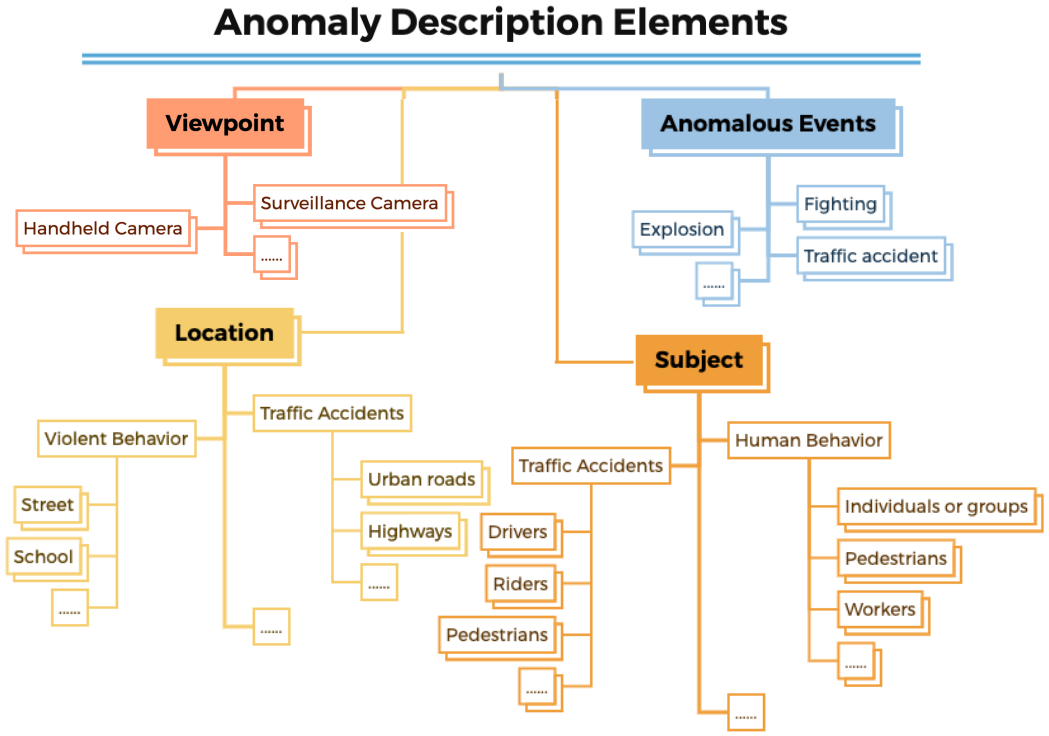}
    \caption{The illustration of anomaly description elements.}
    \label{fig:Anomaly Description Elements}
    \vspace{-3mm}
\end{figure}
Then, a diffusion model \cite{Cogvideox} is used to generate synthetic videos. Specifically, we input the anomaly video description repository $\mathcal{C}_{va}$ and the normal video description repository $\mathcal{C}_{vn}$ into the diffusion model for conditional video generation:
\begin{equation}
\mathcal{V}_{va}, \mathcal{V}_{vn}=\mathcal{G}\left (  \mathcal{C}_{va}, \mathcal{C}_{vn} \right ),
\end{equation}
where 
$\mathcal{G}(\cdot)$ denotes the diffusion models, and $\mathcal{V}_{va},\mathcal{V}_{vn}$ refer to the generated synthetic videos.

Following synthetic video generation, we use a visual encoder $\Phi$ to extract visual features from both the synthetic and real videos:
\begin{equation}
\mathbf{V}_{va},\mathbf{V}_{vn}=\Phi\left ( \mathcal{V}_{va},\mathcal{V}_{vn} \right ) ,
\end{equation}
\begin{equation}
\mathbf{V}_{ra},\mathbf{V}_{rn}=\Phi\left ( \mathcal{V}_{ra},\mathcal{V}_{rn} \right ) ,
\end{equation}
where $\mathbf{V}_{va}=\left \{ \mathbf{V}_{va}^i\right \}_{i=1}^m$ and $\mathbf{V}_{vn}=\left \{ \mathbf{V}_{vn}^i\right \}_{i=1}^m $ represent the synthetic anomalous and normal video sets, while $\mathbf{V}_{ra}=\left \{ \mathbf{V}_{ra}^i\right \}_{i=1}^n$ and $\mathbf{V}_{rn}=\left \{ \mathbf{V}_{rn}^i\right \}_{i=1}^n$ denote the real anomalous and normal video sets.
Each element in visual feature sets represents a visual feature sequence.

For mixed data training,
we combine visual features from synthetic videos with those from real videos in public video anomaly detection datasets.
The mixed anomaly feature sets $\mathbf{V}_{a}$ and normal feature sets $\mathbf{V}_{n}$ are expressed as follows:
\begin{equation}
\mathbf{V}_{a}=\mathbf{V}_{va} \cup \mathbf{V}_{ra},
\end{equation}
\begin{equation}
\mathbf{V}_{n}=\mathbf{V}_{vn} \cup \mathbf{V}_{rn}.
\end{equation}
The resulting feature sets are used to train video anomaly detector with multi-instance learning (MIL).
This hybrid training approach combines complementary scenes and abnormal patterns from both synthetic and real videos, improving the robustness of the video anomaly detector.

\subsection{Synthetic Sample Loss Scaling}
\label{ssec:subhead}

Noting that there exists a domain gap between synthetic and real videos. 
In real-world applications, the data distribution is closer to real videos than to synthetic ones. To simultaneously learn anomaly patterns from synthetic videos while preserving robustness in real videos, we propose a synthetic sample loss scaling (SSLS) strategy.
%

Based on video-level anomaly labels $y$ ($0$ for normal and $1$ for anomalous videos), we introduce an additional data source label $y_s$ ($0$ for real and $1$ for synthetic videos) to distinguish samples during training.
Then, a new scaling factor $\lambda$ is applied to control the impact of synthetic samples. For each sample $i$, the adjusted loss $\mathcal{L}'_i$ is defined as:
\begin{equation}
\mathcal{L}'_i = \left\{
\begin{array}{ll}
\mathcal{L}_i & \text{if \ } y_s[i] = 0 \\
\lambda \cdot \mathcal{L}_i & \text{if \ } y_s[i] = 1
\end{array}
\right. ,
\end{equation}
where $\mathcal{L}_i$ is the original loss for sample $i$.
This strategy balances the contributions of real and synthetic samples, enabling the model to benefit from diverse synthetic videos while reducing the risk of overfitting synthetic video domain. 
The training objective $\mathcal{L}_{total}$ of all samples is computed by summing all adjusted losses:
\begin{equation}
\mathcal{L}_{total}=\sum_{i=1}^{N}L'_i .
\end{equation}

\subsection{Model Training}
\label{ssec:Model training}

Our proposed framework can be applied to most VAD models. In this work, we adopt the recent LAP method \cite{LAP} to train the anomaly detector. 
LAP follows the multi-instance learning (MIL) framework, adopting a Top-k strategy to select the most anomalous segments. These segments are optimized using a binary cross-entropy loss function $\mathcal{L}_{MIL}$:
\begin{equation}
\mathcal{L}_{MIL} =\sum_{y \in [y_a, y_n]}  - \left( y \log(\hat{y}) + (1 - y) \log(1 - \hat{y}) \right) ,
\end{equation}
where $\hat{y}$ is the predicted anomaly score computed by averaging the top-k highest segment-level scores.

The multi-prompt learning loss and pseudo anomaly loss in LAP \cite{LAP} are also employed. Here we denote the additional loss terms from LAP as $\mathcal{L}_{LAP}$.
The final training objective of sample $i$ is defined as:
\begin{equation}
\mathcal{L}_i = \mathcal{L}_{MIL}+\mathcal{L}_{LAP}.
\end{equation}

\section{Experiments}
\label{sec:pagestyle}

\begin{figure}[t]
    \centering    \includegraphics[width=\linewidth]{
    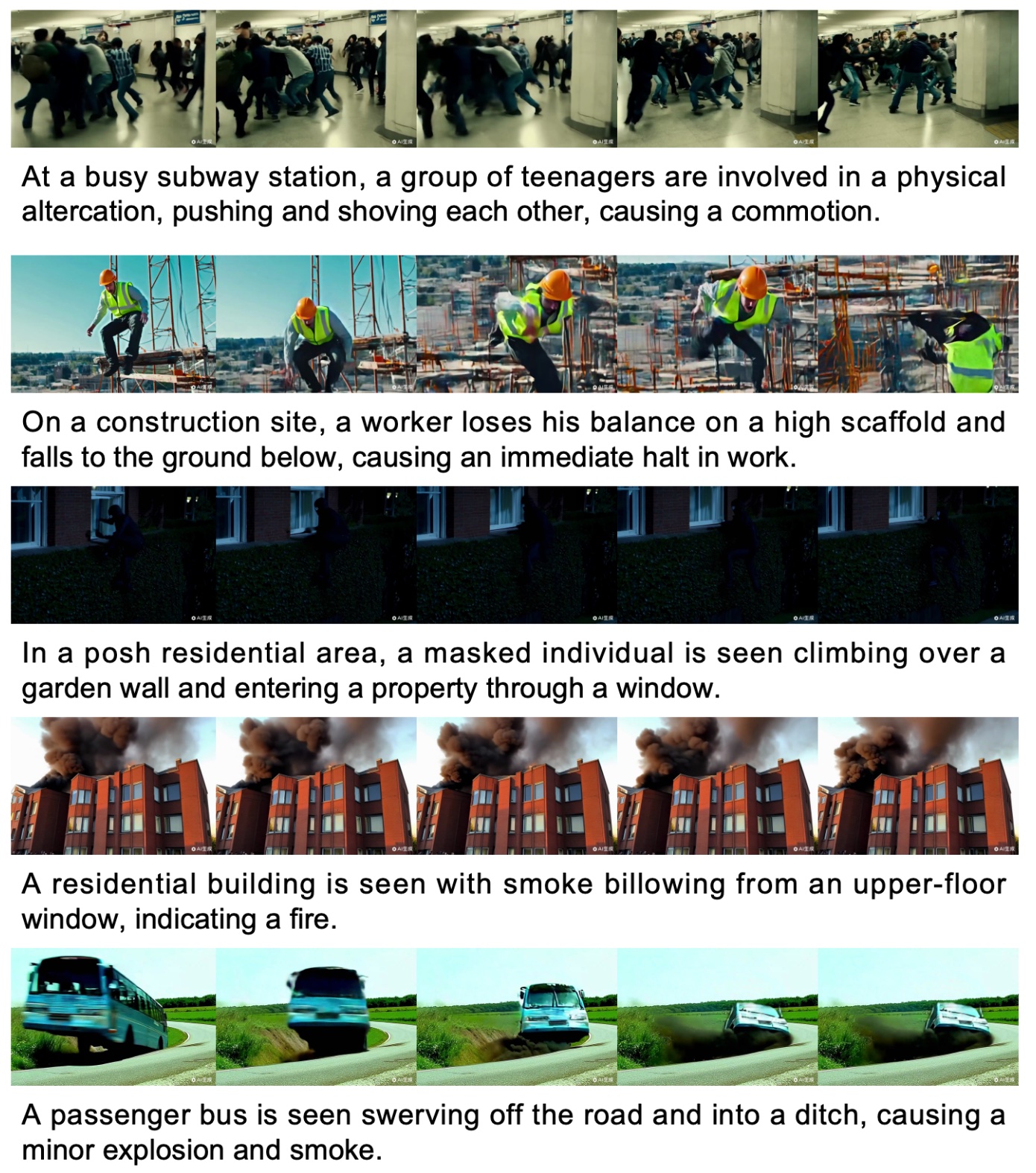}
    \caption{Examples of generated abnormal videos.}
    \label{fig:GVE}
\end{figure}

\begin{table}[t] 
\caption{Frame-level AUC results on UCF-Crime dataset.}
\label{tab: ucfsota}
\centering  
\resizebox{\linewidth}{!}{
\begin{tabular}{c|cccc}
\Xhline{1.8pt} 
Type&Method&Source&Feature&AUC(\%)\\
\Xhline{1.8pt} 
\multirow{2}{*}{\rotatebox{90}{Semi}}
&GODS\cite{wang2019gods}&ICCV'19&I3D&70.5\\
&GCL\cite{zaheer2022generative}&CVPR'22& ResNet&74.2\\
\Xhline{0.5pt} 
\multirow{14}{*}{{\rotatebox{90}{Weakly}}}
&Sultani et al.\cite{sultani2018real}&CVPR'18&I3D&82.1\\
&RTFM\cite{tian2021weakly}&ICCV'21&I3D&84.3\\
&MSL\cite{li2022self}&AAAI'22& VideoSwin&85.6\\
&TEVAD\cite{TEVAD}&CVPR'23&I3D&84.9\\
&MGFN\cite{MGFN}&AAAI'23&VideoSwin&86.6\\
&UR-DMU\cite{UR-DMU}&AAAI'23&I3D&86.9\\
&CLIP-TSA\cite{joo2023clip_tsa}&ICIP'23& CLIP&87.6\\
&SSA\cite{fan2024sneppetAttention}&T-CSVT'24& I3D&86.2\\
&OVVAD\cite{OVVAD}&CVPR'24& I3D&86.4\\
&PEL\cite{PEL}&TIP'24& I3D&86.8\\
&VadCLIP\cite{wu2023vadclip}&AAAI'24& CLIP&88.0\\ 
\cline{2-5}
&RTFM\cite{tian2021weakly}&ICCV'21&CLIP&87.1\\
&\cellcolor{gray!20}GV-VAD(RTFM)&\cellcolor{gray!20}Ours&\cellcolor{gray!20}CLIP&\cellcolor{gray!20}87.3\\
\cline{2-5}
&TEVAD\cite{TEVAD}&CVPR'23&CLIP&87.6\\
&\cellcolor{gray!20}GV-VAD (TEVAD) & \cellcolor{gray!20}Ours & \cellcolor{gray!20}CLIP & \cellcolor{gray!20}88.1 \\
\cline{2-5}
&LAP\cite{LAP}&arxiv'24& CLIP &88.9\\
& \cellcolor{gray!20}\textbf{GV-VAD (LAP)} & \cellcolor{gray!20}\textbf{Ours} & \cellcolor{gray!20}\textbf{CLIP} & \cellcolor{gray!20}\textbf{89.3} \\
\Xhline{1.8pt} 
\end{tabular}
}
\vspace{-3mm}
\end{table}

\subsection{Implementation Details}
The performance is evaluated on UCF-Crime \cite{sultani2018real}, which is a large-scale video anomaly detection dataset, comprising 128 hours of surveillance video from 13 different anomaly categories. 
GPT-4o is employed to construct video descriptions, and CogVideoX \cite{Cogvideox} is used to generate a total of 600 synthetic videos. The number $m$ of generated normal and abnormal videos is 300 each. 
Generated abnormal video samples are shown in \ref{fig:GVE}.

Following previous works, input videos are uniformly segmented into 16-frame clips for visual feature extraction. CLIP-L is used as visual encoder, where the dimension of visual features and semantic features is 768. Adam optimizer is used with learning rate 0.001 and weight decay 0.005. The virtual sample loss scaling factor $\lambda$ defaults to 0.5. 

\subsection{Quantitative Comparison}

The quantitative comparison results of frame-level AUC are presented in Table \ref{tab: ucfsota}. 
We compare our method with prevailing semi-supervised methods \cite{wang2019gods,zaheer2022generative} and weakly-supervised methods \cite{sapkota2022bayesian,zhang2022weakly,li2022self,NG_MIL,TEVAD,COMO,MGFN,UR-DMU,joo2023clip_tsa,fan2024sneppetAttention,OVVAD,PEL,wu2023vadclip,LAP}. 
It can be seen that, compared to state-of-the-art methods, our method achieves the best AUC performance on the UCF crime dataset \cite{sultani2018real}, thanks to the generated videos and the synthetic sample loss scaling.
Moreover, the implementation on different weakly-supervised methods (TEVAD \cite{TEVAD} and RTFM \cite{RTFM}) further demonstrates the robustness and compatibility of our GV-VAD framework.

\subsection{Ablation Study}
To verify the effectiveness of the steps of video generation (VG), video filtering (VF) and synthetic sample loss scaling (SSLS) in GV-VAD, a module ablation experiment is tested on the UCF-Crime dataset, which are shown in Table \ref{tab: module}. It can be seen that the progressive incorporation of VG, VF, and SSLS provide steady AUC performance improvement.

In Table \ref{tab: lambda}, another ablation study is conducted on the synthetic sample loss scaling (SSLS) strategy.
When $\lambda = 1.0$, SSLS is actually disabled, i.e. synthetic data contribute equally as real data. 
Lowering $\lambda$ reduces the influence of synthetic samples during training, helping to focus on learning on real data. 
Thus, moderate scaling (e.g., $\lambda = 0.5$) improves detection performance, while overly small values (e.g., $\lambda = 0.1$) weaken the benefits of synthetic data. 
Moreover, setting $\lambda$ as a learnable parameter yields suboptimal performance.

\begin{table}[t]
    \renewcommand\arraystretch{1}
    \centering
    \caption{Ablation study of proposed modules on UCF-Crime.} 
    \label{tab: module}
    \resizebox{\linewidth}{!}{
    \begin{tabular}{cccIc}
    \Xhline{1.8pt}
        Video Generation & Video Filtering & SSLS & AUC(\%) \\ 
        \Xhline{1.8pt}
        -- & -- & -- & 88.7 \\ 
        \Checkmark & -- & -- & 88.9 \\ 
        \Checkmark & \Checkmark & -- & 89.0 \\ 
        \Checkmark & -- & \Checkmark & 89.0 \\ 
        \Checkmark & \Checkmark & \Checkmark & 89.3 \\ \Xhline{1.8pt}
    \end{tabular}
    }
\end{table}

\begin{table}[t]
    \centering
    \caption{Ablation study of paramenter $\lambda$ in SSLS.}  
    \label{tab: lambda}
    \begin{tabular}{ccccccc}
    \Xhline{1.0pt}
        $\lambda$ & 0.1 & 0.25 & 0.5 & 1.0 & 2.0  & Learnable \\ \Xhline{0.5pt}
        AUC(\%) & 88.7 & 88.9 & 89.3 & 89.0 & 88.6  & 89.2\\ \Xhline{1.0pt}
    \end{tabular}
    \vspace{-3mm}
\end{table}

\begin{figure}[t]
    \centering
    \includegraphics[width=\linewidth]{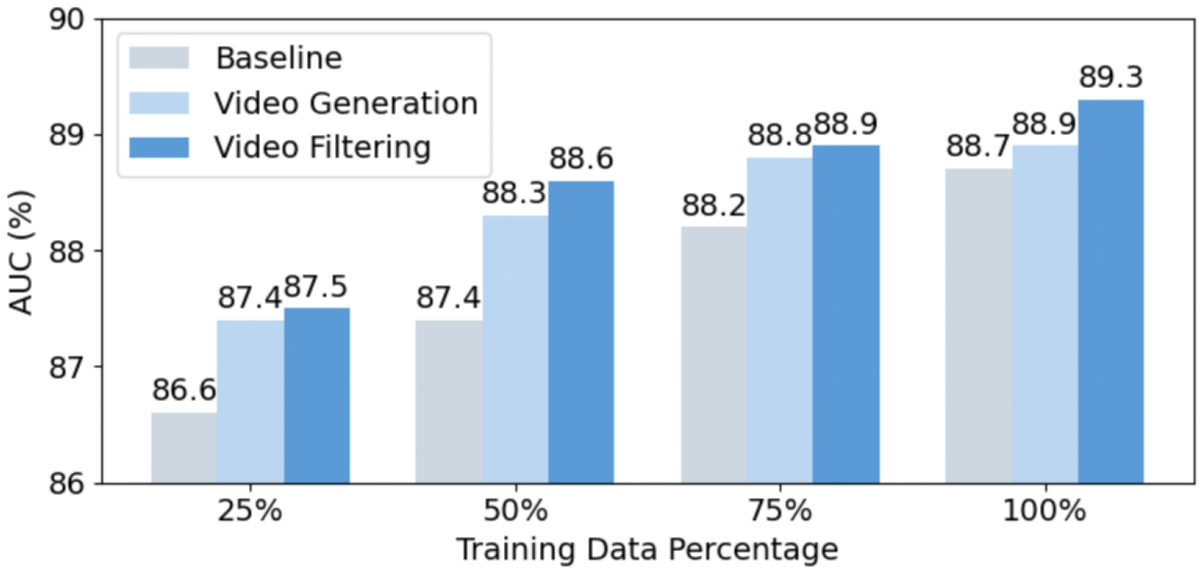}
    \caption{Ablation study on the effect of adding generated data with different amounts of training videos on the UCF-Crime dataset. Best viewed in color.}
    \label{fig:DSR}
\end{figure}

To evaluate the effectiveness of various training data scales, the third ablation study is illustrated in Fig. \ref{fig:DSR}. 
The results on the UCF-Crime dataset show that adding synthetic videos consistently improves performance. Without synthetic data, AUC scores ranged from 86.6\% to 88.7\% across different data scales. Adding generated videos, the AUC at 25\% data scale is increased by 0.9\% to 87.5\%, which is higher than the baseline at 50\% data scale.
Using carefully selected synthetic data further boosted performance to 89.3\% AUC at full data scale, a 0.6\% improvement over baseline.
This shows that our method is valuable for low-resource scenarios with scarce anomaly samples.
\begin{figure}[t]
    \centering
    \includegraphics[width=\linewidth]{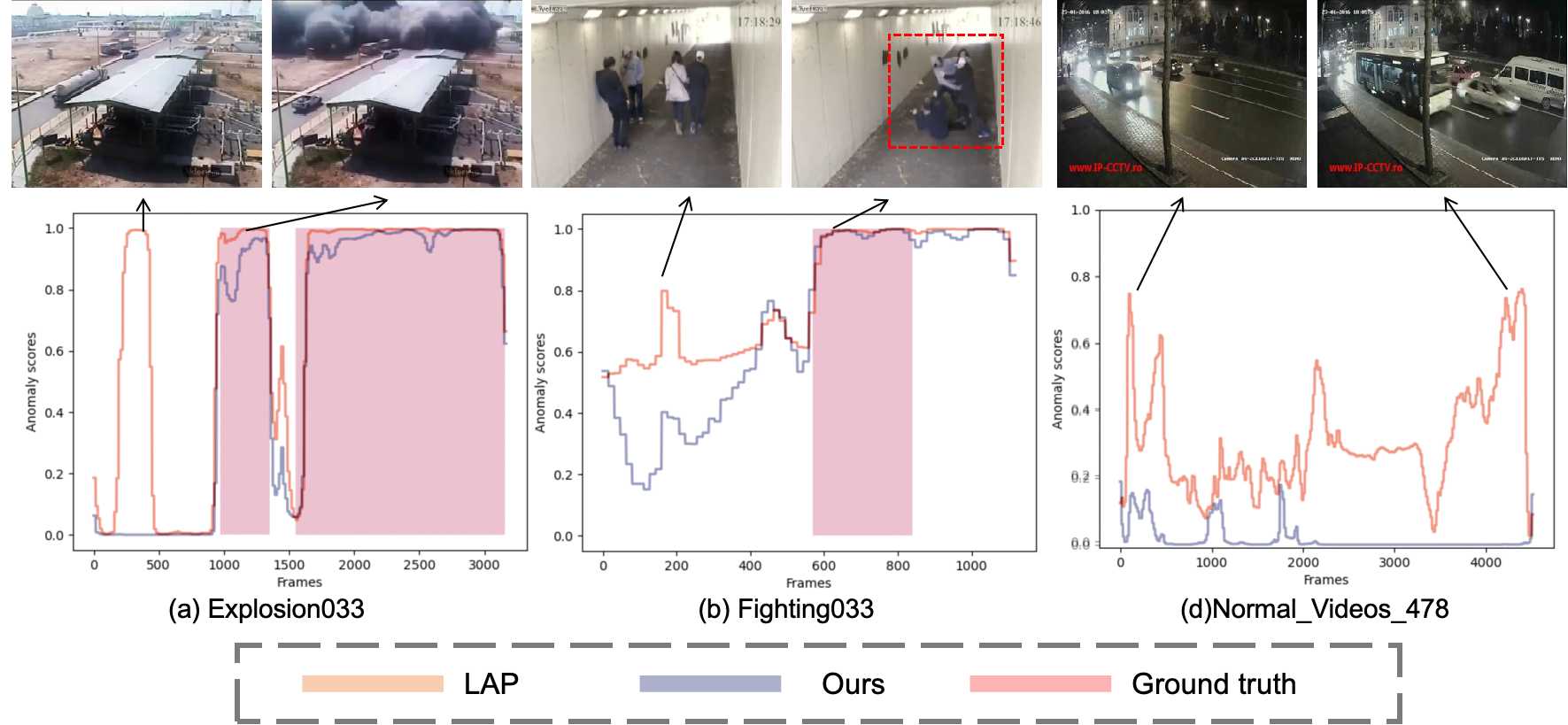}
    \caption{Qualitative comparison of our GV-VAD (blue) and LAP (orange) on the UCF-Crime dataset. The red area corresponds to the abnormal regions of the ground truth. 
    }
    \label{fig:curve}
\end{figure}

\subsection{Qualitative Results and Analysis}

The visual examples of anomaly score plots generated by the proposed GV-VAD are presented in Fig. \ref{fig:curve}. Two anomalous videos and one normal video are presented for illustration. 
In the explosion event (Fig. \ref{fig:curve}(a)) and fighting event (Fig. \ref{fig:curve}(b)), the baseline method triggers false alarms in normal frames.
In the normal video (Fig. \ref{fig:curve}(c)), due to poor lighting conditions, the baseline method still assigns high anomaly scores.
Thus, it cannot accurately discriminate the normal and anomalous regions. 
In contrast, GV-VAD provides more accurate and temporally consistent anomaly predictions, demonstrating enhanced robustness to complex scenes and visual noise.

\section{Conclusion}
\label{sec:typestyle}
In this work, a novel generative-video enhanced weakly-supervised video anomaly detection framework has been proposed to augment training data at low cost. 
A diffusion model-based video generation and filtering method is applied to efficiently expand training data. 
In addition, a synthetic sample loss scaling strategy is utilized to control the influence of synthetic samples for efficient data-mixed training.
Experiments on UCF-Crime demonstrated the effectiveness of the proposed GV-VAD framework.

\begingroup
\small  
\bibliographystyle{IEEEtran}
\bibliography{refs}
\endgroup


\end{document}